\documentclass[sigconf]{acmart}

\usepackage{color}
\usepackage{booktabs}
\usepackage{amsmath}
\usepackage{amsthm}
\usepackage{amsfonts}
\usepackage{multirow}
\usepackage{tabularx}

\usepackage{graphicx}
\usepackage{algorithm}
\usepackage{algorithmic}

\usepackage{adjustbox}
\usepackage{balance}
\makeatletter
\g@addto@macro\normalsize{%
  \abovedisplayskip 1pt plus 1pt minus 1pt%
  \belowdisplayskip \abovedisplayskip
  \abovedisplayshortskip 2pt plus1pt  minus1pt%
  \belowdisplayshortskip 2pt plus1pt minus1pt%
}
\g@addto@macro\small{%
  \abovedisplayskip 2pt plus 1pt minus 1pt%
  \belowdisplayskip \abovedisplayskip
  \abovedisplayshortskip 2pt plus1pt  minus1pt%
  \belowdisplayshortskip 2pt plus1pt minus1pt%
}
\g@addto@macro\footnotesize{%
  \abovedisplayskip 2pt plus 1pt minus 1pt%
  \belowdisplayskip \abovedisplayskip
  \abovedisplayshortskip 2pt plus1pt  minus1pt%
  \belowdisplayshortskip 2pt plus1pt minus1pt%
}
\makeatother

\newcommand{\gengbinzong}[1]{\textcolor{blue}{[Geng: #1]}}

\copyrightyear{2021}
\acmYear{2021}
\setcopyright{acmcopyright}\acmConference[SIGIR '21]{Proceedings of the 44th International ACM SIGIR Conference on Research and Development in Information Retrieval}{July 11--15, 2021}{Virtual Event, Canada}
\acmBooktitle{Proceedings of the 44th International ACM SIGIR Conference on Research and Development in Information Retrieval (SIGIR '21), July 11--15, 2021, Virtual Event, Canada}
\acmPrice{15.00}
\acmDOI{10.1145/3404835.3462902}
\acmISBN{978-1-4503-8037-9/21/07}

\begin{document}
\fancyhead{}


\title{Iterative Network Pruning with Uncertainty Regularization \\ for Lifelong Sentiment Classification}

\author{Binzong Geng$^{1,2*}$, Min Yang$^{2\dagger}$, Fajie Yuan$^{3,6*}$, Shupeng Wang$^{2}$, Xiang Ao$^{4}$, Ruifeng Xu$^{5}$}

\affiliation{\institution{$^{1}$University of Science and Technology of China}}

\affiliation{\institution{$^{2}$Shenzhen Institutes of Advanced Technology, Chinese Academy of Sciences}}

\affiliation{\institution{$^{3}$Westlake University}}

\affiliation{\institution{$^{4}$Key Lab of Intelligent Information Processing of Chinese Academy of Sciences, Institute of Computing Technology, CAS}}


\affiliation{\institution{$^{5}$Harbin Institute of Technology (Shenzhen) \qquad  $^{6}$Tencent}}



\email{{bz.geng, min.yang, wang.sp}@siat.ac.cn,  yuanfajie@westlake.edu.cn, aoxiang@ict.ac.cn, 	xuruifeng@hit.edu.cn}

\thanks{$^{*}$This work was done when Binzong Geng interned at SIAT, Chinese Academy of Sciences. This work was done when Fajie Yuan worked at Tencent (past affiliation) and Westlake University (current affiliation).
}
\thanks{$^{\dagger}$Min Yang is corresponding author.}


\begin{abstract}
Lifelong learning capabilities are crucial for sentiment classifiers to process continuous streams of opinioned  information on the Web. 
However, performing lifelong learning is non-trivial for deep neural networks  as  continually training of incrementally available information inevitably results in catastrophic forgetting or interference. In this paper, we propose a novel \underline{\textbf{i}}terative network \underline{\textbf{p}}runing with uncertainty \underline{\textbf{r}}egularization method for \underline{\textbf{l}}ifelong \underline{\textbf{s}}entiment classification (IPRLS), which leverages the principles of network pruning and weight regularization.
By performing network pruning with uncertainty regularization in an iterative manner, IPRLS can adapt a single BERT model to work with continuously arriving data from multiple domains while avoiding catastrophic forgetting and interference. 
Specifically, we leverage an iterative pruning method to remove redundant parameters in large deep networks so that the freed-up space can then be employed to learn new tasks, tackling the catastrophic forgetting problem.  Instead of keeping the old-tasks fixed when learning new tasks, we also use an uncertainty regularization based on the Bayesian online learning framework to constrain the update of old tasks weights in BERT,  which enables positive backward transfer, i.e. learning new tasks improves performance on past tasks while protecting  old knowledge from being lost. 
In addition, we propose a task-specific low-dimensional residual function in parallel to each layer of BERT, which makes IPRLS less prone to losing  the knowledge saved in the base BERT network when learning a new task.
Extensive experiments on 16 popular review corpora demonstrate that the proposed IPRLS method significantly outperforms the strong baselines for lifelong sentiment classification. 
For reproducibility, we submit the code and data at: \textcolor{blue}{\url{https://github.com/siat-nlp/IPRLS}}.
\end{abstract}



\maketitle

\section{Introduction}
With the increase of large collections of opinion-rich documents on the Web, much focus has been given to sentiment classification that targets at automatically predicting the sentiment polarity of given text. In recent years, deep learning has achieved great success and been almost dominant in the field of sentiment classification \cite{tai2015improved,wang2016dimensional,wen2019memristive}. Powerful deep neural networks 
have to depend on large amounts of annotated
training resources. However, labeling large datasets is usually time-consuming and labor-intensive, creating significant barriers when applying the trained sentiment classifier to new domains. In addition, no matter how much data is collected and used to train a sentiment classifier, it is difficult to cover all possible domains of opinioned data on the Web. Thus, when deployed in practice, a well-trained sentiment classifier often performs unsatisfactorily. 

A sentiment classifier operating in a production environment  often encounters
 continuous streams of information~\cite{parisi2019continual} and thereby is required to expand its knowledge to new domains\footnote{Note that by a task, we mean a category of products. In this paper, we use the terms domain and task interchangeably, because each of our tasks 
is from a distinct domain.}.
The ability to continually learn over time by grasping new knowledge and remembering  previously learned experiences is referred to as lifelong or continual learning \cite{parisi2019continual}.
Recently, there are several studies \cite{chen2015lifelong,wang2018lifelong,lv2019sentiment} that utilize lifelong learning to boost the performance of sentiment classification in a changing environment.  
\citeauthor{chen2015lifelong} \shortcite{chen2015lifelong} proposed a lifelong learning method for sentiment classification based on Naive Bayesian framework and stochastic gradient descent.
\citeauthor{lv2019sentiment} \shortcite{lv2019sentiment} extended the work of \cite{chen2015lifelong} with a neural network based approach.
However, the performances of previous lifelong sentiment classification techniques are far from satisfactory. 

Lifelong learning has been a long-standing challenge for deep neural networks. It is difficult to automatically balance the trade-off between stability and plasticity in lifelong learning \cite{mermillod2013stability,ahn2019uncertainty}.  On the one hand, a sentiment classifier is expected to reuse previously acquired knowledge, but focusing too much on stability may hinder the classifier from quickly adapting to new tasks. 
On the other hand, when the classifier pays too much attention to 
plasticity, it may quickly forget previously-acquired abilities. 
One possible solution is to efficiently reuse previously acquired knowledge when processing new tasks, meanwhile avoiding forgetting previously-acquired abilities. That is, on the one hand, consolidated knowledge is preserved to hold the long-term durability and prevent catastrophic forgetting when learning new tasks over time. On the other hand, in certain cases such as immersive long-term experiences, old knowledge can be 
modified or substituted to refine new knowledge and avoid knowledge interference~\cite{parisi2019continual}. 

In this paper, we propose a novel \underline{\textbf{i}}terative network \underline{\textbf{p}}runing with uncertainty \underline{\textbf{r}}egularization method for \underline{\textbf{l}}ifelong \underline{\textbf{s}}entiment classification (IPRLS).  IPRLS deploys BERT ~\cite{devlin2019bert} as the base model for sentiment classification given its superior performance in the recent literature. To resolve the stability-plasticity dilemma \cite{ahn2019uncertainty}, we leverage the principles of network pruning and weight regularization, sequentially integrating the important knowledge from multiple sequential tasks into a single BERT model while ensuring minimal decrease in accuracy. 
Specifically, in each round of pruning, we use a weight-based pruning technique to free up a certain fraction of eligible weights from each layer of BERT after it has been trained for a task, and the released parameters can be modified for learning subsequent new tasks. Instead of keeping the old-tasks weights fixed when learning new tasks as in previous works \cite{rusu2016progressive,mallya2018packnet}, we incorporate an uncertainty regularization based on the Bayesian online learning framework into the iterative pruning procedure. The uncertainty regularization constrains the update of the old-task weights in BERT, which guides the model to gracefully update the old-task weights and enables positive backward transfer. Since a limited architecture cannot ensure to remember the knowledge incrementally learned from unlimited tasks, we also allow expanding the architecture to some extent by adding a task-specific low-dimensional residual ``in parallel'' with each BERT layer to learn important knowledge for each new task.

This paper has three main contributions listed as follows:
\begin{itemize}
    \item We propose a novel iterative network pruning with uncertainty regularization approach for lifelong sentiment classification, which alleviates the catastrophic forgetting issue while overwriting the old knowledge in favor of the acquisition and refinement of new knowledge. Our model takes full advantage of both architecture-based and regularization-based lifelong learning methods.  
    \item We introduce a task-specific low-dimensional multi-head attention layer that is added in parallel to each layer of the base BERT model, which can further retain the important knowledge for the corresponding new task. It achieves better and more stable results than previous methods, with only a small fraction of additional task-specific parameters.
    \item We conduct extensive experiments on 16 popular review datasets. The experimental results demonstrate that our IPRLS approach outperforms strong baseline methods significantly for lifelong sentiment classification.
\end{itemize}

\section{Related Work}
\subsection{Sentiment Classification}
Sentiment classification has become a research hotspot, which has attracted much attention from both academia and industry. 
Recently, pre-trained language models, such as GPT \cite{radford2018improving}, BERT \cite{devlin2019bert}, RoBERTa \cite{liu2019roberta}, ALBERT \cite{lan2020albert},  have been proposed and applied to many NLP tasks (including sentiment classification), yielding state-of-the-art performances. 
While some studies claim that pre-trained language models are ``universal'' for various NLP tasks, there is no theoretical justification for this. In addition, based on empirical evidence, the performance of  fine-tuned BERT deteriorates substantially when dealing with new domains or new tasks. This motivates many researchers to study sentiment analysis in changing domains. 

To better deal with new domain data, plenty of methods have been proposed for cross-domain sentiment classification, which train
sentiment classifiers on  labeled 
source domain data so as to generalize to new domains. 
Among them, structural correspondence learning (SCL) \cite{blitzer2006domain} is a representative domain adaptation approach, by inducing correspondences among features from different domains.
To make use of deep learning, several neural network models were developed to learn common features or shared parameters for sentiment classification in the domain adaption scenario, and these models have shown impressive performance.
Stacked denoising auto-encoder (SDA) \cite{glorot2011domain} and its variants have become representative domain adaptation approachesb based on deep learning, which automatically learn domain-agnostic features by training data from distinct domains.  Subsequently, \cite{ziser2018pivot} proposed  combining the denoising auto-encoder and conventional pivot-based methods so as to further boost the accuracy.
Compared to single-source domain adaptation, multi-source domain adaptation (MDA) is not only feasible in practice but also valuable in performance \cite{wu2016sentiment,guo2018multi,du2020adversarial}. 

However, in cross-domain sentiment classification scenario, the features relevant to the new domains are learned through modification of the network weights, thus the important weights for prior domains might be altered, which may lead to performance deterioration of the prior domains. Continual learning is a prominent technique to tackle the ``catastrophic forgetting'' \cite{kirkpatrick2017overcoming} issue.

\subsection{Lifelong and Continual Learning}
Continual learning, also known as lifelong learning or incremental learning, sequentially learns a sequence of tasks \cite{aljundi2019task,ahn2019uncertainty}. 
Some progress has been achieved for lifelong/continual learning  in domains, such as computer vision  \cite{aljundi2019task,Kirkpatrick3521}, recommender systems~\cite{yuan2020parameter,yuan2020one} and reinforcement learning \cite{mermillod2013stability}. 
Generally, these continual learning techniques can be divided into three categories: memory-based approaches \cite{lopez2017gradient}, architecture-based approaches \cite{rusu2016progressive}, and regularization-based approaches \cite{kirkpatrick2017overcoming}.  
In this paper, we attempt to improve the performance of continual learning by taking advantage of both architecture-based and  regularization-based techniques.

Architecture-based methods allocate a dedicated capacity inside a model for each task~\cite{sun2019lamol}. The corresponding parameters are frozen and may not be changed after training a specific task.
The progressive neural network \cite{rusu2016progressive} was a representative architecture-based method, which replicated the network architecture for each new task.
PathNet \cite{fern2017pathnet} utilized an evolutionary method to find an optimal path through a huge fixed-size neural network for each task, and only the selected path is trainable for the corresponding task. 
PackNet \cite{mallya2018packnet} employed a weight-based pruning method to free up redundant parameters of a deep network, which could then be used to learn new tasks.
However, the complexity of the model architecture increases as the number of tasks grows, making the deep model difficult to train.

Regularization-based methods add penalty terms to minimize deviation from previously learned parameters while updating specific weights for new tasks. For example, 
\cite{kirkpatrick2017overcoming} proposed elastic weight consolidation (EWC), which estimates parameter importance by calculating a Fisher information matrix and assign high penalty to important parameters.
\cite{schwarz2018progress} proposed online EWC to accumulate the importance of the task steam, instead of tracking the importance of parameters per task. 
\cite{ahn2019uncertainty}
proposed node-wise uncertainty based on the Bayesian online learning framework and devised uncertainty regularization terms for dealing with the stability-plasticity dilemma.  

In parallel, there are several works devoted to improving 
the performance of sentiment classification in a continual learning scenario \cite{chen2015lifelong,wang2018lifelong,lv2019sentiment}. For example, \cite{chen2015lifelong} proposed a lifelong sentiment classification technique, which explored a Bayesian optimization framework and exploited Knowledge via penalty terms.
\cite{lv2019sentiment} extended the work of \cite{chen2015lifelong} with a neural network based approach.

Different from previous works, we propose a novel BERT-based model to learn multiple tasks sequentially by taking advantage of both architecture-based and regularization-based continual learning methods. In addition, we also introduce a task-specific low-dimensional multi-head attention layer that is added in parallel to each layer of the base BERT model, which can further retain the important knowledge for each new task.

\section{Our Methodology}
\subsection{Task Definition}
Suppose the sentiment classifier $\mathcal{M}_{1:k}$ has performed learning on a sequence of $k$ tasks from 1 to $k$, denoted as  $\mathcal{T}_{1:k}=\{\mathcal{T}_1, \cdots, \mathcal{T}_k\}$. The goal of lifelong sentiment classification is to use the knowledge gained in the past $k$ tasks to help learn a better classifier $\mathcal{M}_{1:k+1}$ for the $k+1$-th task $\mathcal{T}_{k+1}$  while avoid forgetting knowledge learned from past tasks. In this paper, each task is a sentiment classification problem for a specific domain,  aiming to classify the reviews as positive or negative. We use the terms ``domain'' and ``task'' interchangeably, because each of our tasks is from a different domain. 

\begin{figure*}
    \centering
    \includegraphics[scale=0.7]{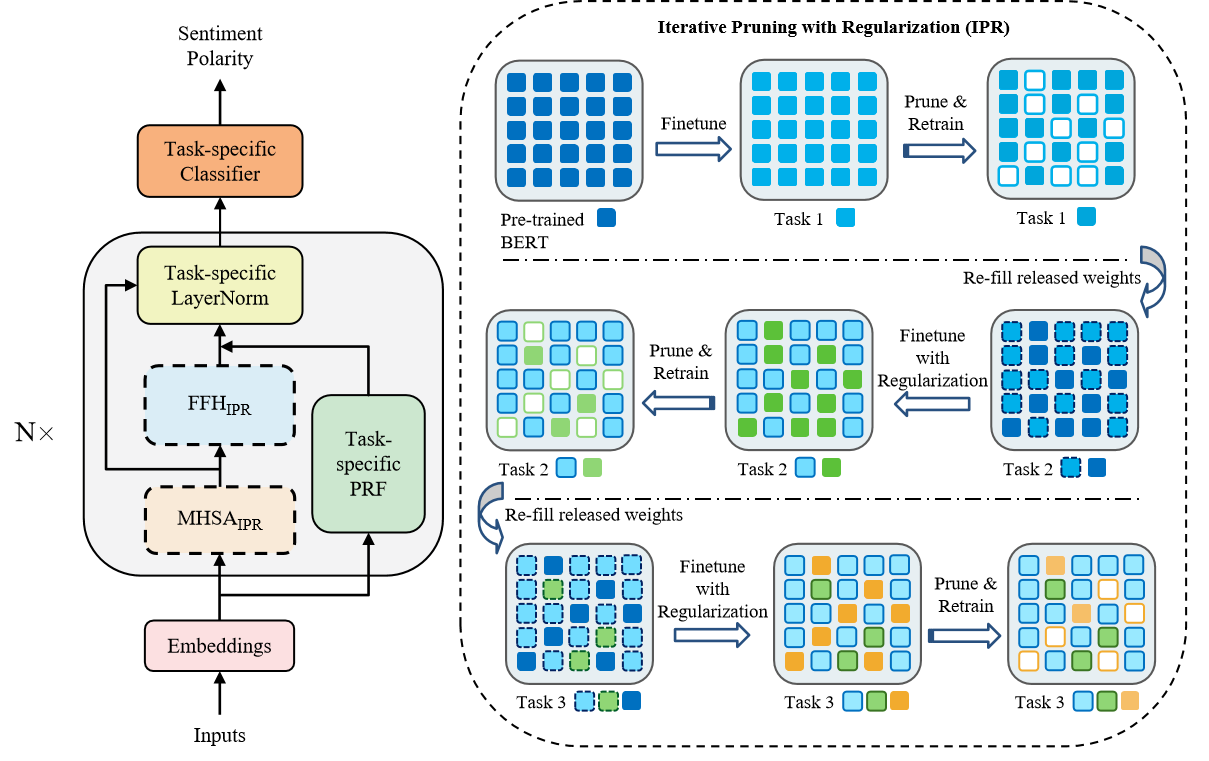}
    \caption{Illustration of the overall architecture based on the BERT model (left) and the processing flow of the iterative pruning with uncertainty regularization method (right).} 
    \label{fig:IPRLS}
\end{figure*}

\subsection{Overview}
In this paper, we use BERT as the base model to build the sentiment classifier. BERT is an essential representative of the rapidly developing pre-trained models, which shows superior performance for various NLP tasks. In general, when the tasks arrive sequentially, BERT would suffer from ``catastrophic forgetting'' on old tasks while learning the new ones. To alleviate the ``catastrophic forgetting'' issue, we take advantage of both architecture-based and regularization-based continual learning methods to improve the performance of BERT for life-long sentiment classification.
Specifically, we explore two mechanisms to facilitate the BERT model to preserve knowledge important to prior tasks when learning a new task. 
First, we explore an iterative pruning with uncertainty regularization to integrate important knowledge from multiple tasks into a single BERT model while ensuring minimal drop in accuracy. 
Second, we add a task-specific parallel residual function ``in parallel'' with each BERT layer to further retain the most important new knowledge while adapting to a new task.
Next, we will elaborate on BERT, the iterative pruning with uncertainty regularization, and the task-specific parallel residual function. 


\subsection{Preliminary: The BERT Model}
The well-known property of the pre-trained BERT model is that it can be  fine-tuned to create state-of-the-art models for a variety of downstream tasks without substantial task-specific architecture modifications~\cite{devlin2019bert}. 
In particular, the architecture of the BERT model is a multi-layer bidirectional Transformer encoder, which is made up of a stack of $M$ identical layers. Each layer has two sub-layers, in which the first layer is a multi-head self-attention layer and the second is a standard fully connected feed-forward layer. The residual connection is applied to each of the two continuous sub-layers, followed by layer normalization~\cite{ba2016layer}. 
The multi-head attention layer is the core of the transformer architecture that transforms the hidden state for each element of a sequence based on the others.  
Formally, the $i$-th attention ``head'' can be computed as:
\begin{equation}
\label{eq:attention}
\small
  {\rm Attention}_i (\mathbf{h}_j) = \sum_{t=1}^{L} {\rm softmax} \left(\frac{W^q_i\mathbf{h}_j \cdot W^k_i \mathbf{h}_t}{\sqrt{d/n}}\right)W^v_i \mathbf{h}_t
\end{equation}
where $\mathbf{h}_j$ or $\mathbf{h}_t$ is a hidden vector for particular sequence element. $L$ represents the sequence length. $W^q_i$, $W^k_i$, $W^v_i$  are projection parameters. $d$ represents the dimension of $\mathbf{h}_j$ and $n$ is the number of attention heads. The multi-head self-attention layer, which we write as $\mathbf{MHA}(\cdot)$, consists of $n$ different dot-product attention mechanisms.
We compute the multi-head attention   $\mathbf{MHA}(\cdot)$ by concatenating the $n$ attention heads via linear transformation:
\begin{equation}
\label{eq:MH}
\small
  {\rm \mathbf{MHA}}(\mathbf{h}) = W^c [{\rm Attention}_1 (\mathbf{h}), \ldots, {\rm Attention}_n (\mathbf{h})]
\end{equation}
where  $W^c$ is the projection parameter. 
The residual connection is applied to each of the two continuous sub-layers, followed by layer normalization:
\begin{equation}
\label{eq:MSA}
 {\mathbf{MHAL}}(\mathbf{h}) = {\rm \mathbf{LN}}(\mathbf{h}+{\rm \mathbf{MHA}}(\mathbf{h}))
\end{equation} 
where $\mathbf{LN}(\cdot)$ represents layer normalization. $\mathbf{MHAL}(\cdot)$ denotes the multi-head attention layer after layer normalization.  

For the second sub-layer, the fully connected layer (FFN) is then applied to the multi-head self-attention layer.
We define $\mathbf{FFN}(\cdot)$ as a standard feed-forward network:
\begin{equation}
\label{eq:FFN}
  {\rm \mathbf{FFN}}(\mathbf{h}) = W_2^f g(W_1^f {\rm \mathbf{MHAL}}(\mathbf{h}) + b_1) + b_2
\end{equation}
where $g$ is a non-linearity.  $W_1^f$ and $W_2^f$ are weight matrices in the FFN function. Finally, the layer normalization is applied to the FFN function, forming the BERT layer which we write as $\mathbf{BL}(\cdot)$:
\begin{equation}
\label{eq:bert}
  {\rm \mathbf{BL}}(\mathbf{h}) = {\rm \mathbf{LN}}({\rm \mathbf{MHAL}}(\mathbf{h})+{\rm \mathbf{FFN}}(\mathbf{h}))
\end{equation}
Similar to ~\cite{stickland2019bert}, the entire BERT model of this paper is simply a stack of 12 BERT layers. The final hidden state of the first token (CLS) of every input sequence is used for  transforming to the output \cite{devlin2019bert}. 
The BERT model can be optimized by minimizing the cross-entropy loss $\mathcal{L}_{CE}$ via stochastic gradient descent.  

\subsection{Iterative Pruning with Uncertainty Regularization}
\paragraph{\textbf{Motivation}} 
To alleviate the ``catastrophic forgetting'' issue of the BERT model, a possible way is to preserve the old-task weights that were already learned and expand the network by adding nodes or weights when training new tasks. It is intuitive that there exist (i) generic concepts that characterize reviews across many domains, and (ii) domain-specific features that are highly specific to sub-domain. These generic concepts would likely help to perform transfer across domains.  In most previous architecture-based continual learning methods \cite{rusu2016progressive,mallya2018packnet}, the old-task weights remain fixed, and only the released weights are adapted for new tasks. 
Since the old-task weights remain fixed, it keeps the performance of all learned tasks. 
However, the storage or memory complexity of the model architecture increases as the number of tasks grows, making the deep model difficult to train. 
To alleviate the aforementioned issue, we propose a novel Iterative Pruning with uncertainty Regularization (IPR) method. Instead of keeping the old-tasks fixed when learning new tasks, we incorporate an uncertainty regularization \cite{blundell2015weight,ahn2019uncertainty} based on the Bayesian online learning framework~\cite{opper1998bayesian} into the iterative pruning procedure. The uncertainty regularization constrains the update of the old tasks linear transforms in BERT, which guides the model to update the old-task weights gracefully. 

In particular, this work follows a continual learning scenario. The sentiment classifier has performed learning on a sequence of unlimited tasks. 
We employ an iterative, three-stage framework for lifelong sentiment classification, inspired by previous work \cite{mallya2018packnet}. First, we train an initial network for a new task by applying uncertainty regularization on the whole preserved weights of the BERT model. Second, a weight-based pruning technique is used to free up less important parameters of the BERT model after it has been trained for a new task, with minimal drop in accuracy. Third, we re-train the network after pruning to regain the accuracy for the current task. Next, we will introduce these three modules in detail.

\subsubsection{\textbf{Initial Network Training with Uncertainty Regularization}}
Suppose a compact model that can cope with tasks 1 to $k$ has been built and available, denoted as $\mathcal{M}_{1:k}$.
We denote the model weights preserved for tasks 1 to $k$ as $W_{1:k}^P$. The freed up weights associated with task $k$ are represented as $W_{k}^F$. We let the freed up parameters $W_{k}^F$ come back from zero, which can be modified repeatedly for learning subsequent tasks. 
When dealing with a new task $k+1$, we first train a network $\mathcal{M}_{k+1}$ on the training data for task $k+1$ (initial network training). 
Instead of fixing the preserved weights for old tasks when learning a new task as in \cite{rusu2016progressive,mallya2018packnet}, we exploit an uncertainty regularization on the preserved weight $W_{1:k}^P$ to protect the old knowledge from being lost, inspired by \cite{ahn2019uncertainty}. It is noteworthy that the preserved weights $W_{1:k}^P$ are the linear transforms (i.e., $W^q$, $W^k$, $W^v$, $W^c$, $W^f$) in BERT, defined in Eq. (\ref{eq:attention}),  Eq. (\ref{eq:MH}), and Eq. (\ref{eq:FFN}). We constrain the update of these linear transforms since they play the most important role in BERT for  sentiment classification and take up most of the storage space. 
 
Specifically, the uncertainty regularization is based on the Bayesian online learning framework. For the sake of simplicity, we use $W^{\mathit{UR}}_{1:k}$ to represent the preserved weights $W_{1:k}^P$ with uncertainty regularization. $W^{\mathit{UR}}_{1:k}$ are controlled by two new parameters ($\Phi$, $\boldsymbol{\sigma}$) via Gaussian mean-field approximation: 
\begin{equation}
W^{\mathit{UR}} = \Phi + \boldsymbol{\upsilon} \cdot \boldsymbol{\tau} \cdot \boldsymbol{\sigma}
\end{equation}
where  $\Phi$ represents the mean weight  for preserved parameters $W_{1:k}^P$.  
 $\boldsymbol{\sigma}$ indicates the standard deviation term. $\boldsymbol{\upsilon}$ is a hyperparameter to control the impact of $\boldsymbol{\sigma}$. $\boldsymbol{\tau}$ is a vector, in which each element is a scalar randomly chosen from normal distribution $\mathcal{N}(0,1)$.
The additional parameters ($\Phi$, $\boldsymbol{\sigma}$) can be learned via the standard back-propagation \cite{ahn2019uncertainty}. 

In this paper, we implement three schemes for updating the new parameters from task $k$ to task $k+1$.  
First, we devise a variance-level regularization such that the weight parameters with low uncertainty (i.e., small variance) can get high regularization strength. In the Bayesian online framework, we can easily use variance to correspond to the uncertainty of weight, and low variance means low uncertainty.
The idea behind this scheme is that when any of the 
weights from the last task are significantly updated during the learning of the new task, the current layer will combine the changed information from the lower layer. Thus, the old-task 
weights get significantly modified, hurting the performance of old tasks.
Formally, the weight-level regularization term ${\rm REG}_{1}$ is defined as:
\begin{equation}
\small
  {\rm REG}_{1}(\Phi,\boldsymbol{\sigma}) =   \left\|\max \left\{\frac{\boldsymbol{\sigma}_{\text {init }}^{(l)}}{\boldsymbol{\sigma}_{k}^{(l)}},\frac{\boldsymbol{\sigma}_{\text {init }}^{(l-1)}}{\boldsymbol{\sigma}_{k}^{(l-1)}}\right\} \odot\left(\Phi_{k+1}^{(l)}-\Phi_{k}^{(l)}\right)\right\|_{2}^{2}
\end{equation}
which $\odot$ denotes the element-wise multiplication. $\boldsymbol{\sigma}_{\rm init}^{(l)}$ represents the initial standard deviation hyperparameter for all weights on the $l$-th layer of BERT. $\Phi_{k}^{(l)}$ and $\boldsymbol{\sigma}_{k}^{(l)}$ are the mean and variance weights for layer $l$ and task $k$. $\boldsymbol{\sigma}_{\rm init}^{(l)}$ is set to control the stability of the learning process. 

Second, to consider the magnitude of the learned weights (i.e., $\Phi^{(l)}_{k}$), we devise an additional regularization term ${\rm REG}_{2}$. 
Inspired by \cite{blundell2015weight}, we use the the ratio $(\Phi/\boldsymbol{\sigma})^2$ to judge the the magnitude of weights\footnote{We expand $\boldsymbol{\sigma}$ to have the same dimension as $\Phi$ by repeating the vector $\boldsymbol{\sigma}$ $d_{in}$ times, where $d_{in}$ is the dimension of the input representation.}, which helps to control the stability of the  whole learning process.
Formally, the second regularization term ${\rm REG}_{2}$ is defined as:
\begin{equation}
\small
  {\rm REG}_{2}(\Phi,\boldsymbol{\sigma}) =  \left(\boldsymbol{\sigma}_{\text{init}}^{(l)}\right)^{2}\left\|\left(\frac{\Phi_{k}^{(l)}}{\boldsymbol{\sigma}_{k}^{(l)}}\right)^{2} \odot\left(\Phi_{k+1}^{(l)}-\Phi_{k}^{(l)}\right)\right\|_{1}
\end{equation}
This regularization term tends to promote sparsity. 

Third, as mentioned above, ${\rm REG}_{1}$ and ${\rm REG}_{2}$ are devised to encourage  $\Phi_{k+1}^{(l)}$ to get close to  $\Phi_{k}^{(l)}$. We devise an additional regularization term ${\rm REG}_{3}$ to inflate $\boldsymbol{\sigma}_{k+1}$ to get close to $\boldsymbol{\sigma}_{k}$, which can further prevent the catastrophic forgetting: 
\begin{equation}
{\rm REG}_{3}(\boldsymbol{\sigma}) = \left(\frac{\boldsymbol{\sigma}_{k+1}^{(l)}}{\boldsymbol{\sigma}_{k}^{(l)}}\right)^{2}-\log \left(\frac{\boldsymbol{\sigma}_{k+1}^{(l)}}{\boldsymbol{\sigma}_{k}^{(l)}}\right)^{2}
\end{equation}

Finally, we combine the three regularization terms to form the overall uncertainty regularization $\mathcal{L}_{\mathit{REG}}$:
\begin{equation}
\label{eq:REG}
\small
   \mathcal{L}_{\mathit{REG}} =  \sum_{l=1}^{B}[\frac{\alpha}{2}{\rm REG}_{1}(\Phi,\boldsymbol{\sigma})  +\beta {\rm REG}_{2}(\Phi,\boldsymbol{\sigma}) 
     +\frac{\gamma}{2}{\rm REG}_{3}(\boldsymbol{\sigma})] 
\end{equation}
which $B$ is the number of layers in the network. \textbf{The overall uncertainty regularization objective $\mathcal{L}_{\mathit{REG}}$ is then combined with the cross-entropy loss function  $\mathcal{L}_{\mathit{CE}}$ of the BERT network for the initial network training.}  $\alpha$, $\beta$, and $\gamma$ are hyperparameters which control the importance of the corresponding regularization terms. 

With the uncertainty regularization (three constraints), the old-task 
weights will update cautiously and appropriately when learning a new task.
On the one hand, we can learn more knowledge gained in the past $k$ tasks to help learn a better classifier for the new task (i.e., forward knowledge transfer). 
On the other hand, since all the parameters can update when learning a new task, 
the old tasks can also benefit from the knowledge learned by the new tasks (i.e., backward knowledge transfer).  


\subsubsection{\textbf{Network Pruning}}
We use a weight-based pruning technique to free up a certain fraction of the released weights $W_{k}^F$ across all layers of the BERT model, with minimal deterioration in performance. We let the parameters to be freed up (i.e., $W_{k+1}^F$) associated with task $t+1$ come back from zero, which can be modified repeatedly for learning subsequent tasks. While the remaining parameters are represented as ${W}_{k+1}^P$, which are temporal preserved parameters for task $t+1$. It is noteworthy that, similar to the uncertainty regularization, we solely perform pruning on the linear transforms of the BERT model.

Different from most pruning-based methods that prune the weights according to their absolute magnitude, we exploit a heuristic for pruning network weights learned by variational inference, inspired by \cite{NIPS2011_4329}. Specifically, we sort the weights in each layer by the magnitude of the ratio $\frac{\Phi}{\boldsymbol{\sigma}}$, encouraging the model to preserve the weights that have high absolute magnitude and low uncertainty. 
In each round of pruning, we release a certain percentage\footnote{We free up 40\% of  parameters for the first task and release 75\% of parameters for subsequent tasks.} of eligible weights that have the lowest magnitude of ratio $\frac{\Phi}{\boldsymbol{\sigma}}$. Note that we just perform pruning on the preserved weights that belongs to the current task, and do not update weights belonging to a prior task.

\subsubsection{\textbf{Network Re-training}}
Pruning a network would lead to performance deterioration due to the sudden changes of network connectivity. This is especially obvious when the pruning ratio is higher~\cite{yuan2020one}. To restore the original performance of the network after pruning, we need to re-train the temporal preserved weights $W_{k+1}^P$ for several epochs. After the re-training procedure, we can obtain the overall preserved weights $W_{1:k+1}^P$ for task 1 to $k+1$ by combining $W_{1:k}^P$ and  $W_{k+1}^P$, with minimal loss of performance on prior tasks.

When performing inference for the $k+1$-th task, the freed up weights are masked to ensure that the network state matches the one learned during training, similar to previous work \cite{mallya2018packnet}.  That is, the released weights $W_{k+1}^F$ only have to be masked in a binary on/off fashion during multiplication. This makes the implementation easier by the matrix-matrix multiplication kernels.

The initial network training, network pruning, and network re-training procedures are performed iteratively for learning multiple new tasks, as illustrated in Figure \ref{fig:IPRLS}.  We summarize the overall learning process of iterative network pruning with uncertainty regularization in Algorithm \ref{alg}.

\begin{algorithm}[t]
\begin{algorithmic}
\STATE \textbf{Input}: The sentiment classifier $\mathcal{M}_{1}$ for tasks 1; the released weights $W^F_1$; the preserved weights $W^P_{1}$.
\STATE \textbf{Output}: The sentiment classifier $\mathcal{M}_{K}$ for tasks 1 to $K$.
\end{algorithmic}
	\caption{The learning process of iterative pruning with uncertainty regularization.}
	\label{alg}
	\begin{algorithmic}[1]
    	\FOR{each task $k=2$ to $K$}
    	
    	\STATE Pre-train a initial model for task $k$ by updating released parameters $W^F_{k-1}$ via cross-entropy loss $\mathcal{L}_{\mathit{CE}}$;
    
    	\STATE Update the preserved weights $W_{1:k-1}^P$ for tasks 1 to $k-1$ with uncertainty regularization via $\mathcal{L}_{\mathit{CE}} + \mathcal{L}_{\mathit{REG}}$;
    	
    	\STATE Perform network pruning on released weights $W_{k-1}^F$ and obtain released weights $W_k^F$ for learning subsequent tasks;
    	\STATE Obtain temporal preserved weights $W^P_k$ for task $k$ as $W^P_k = W_{k-1}^F \setminus W^F_{k}$;
    	\STATE Re-train $W^P_k$ for a smaller number of epochs,
    	 and get the overall preserved weights $W_{1:k}^P$ for tasks 1 to $k$ as $W_{1:k}^P = W_{1:k-1}^P \cup W^P_k$. 
    	\ENDFOR
	\end{algorithmic}
	\begin{algorithmic}
	\STATE The preserved weights $W_{1:K}^P$ are used to build sentiment classifier $\mathcal{M}_K$. 	    
	\end{algorithmic}
\end{algorithm}

\subsection{Parallel Residual Function}
The preserved parameters for old tasks are co-used for learning new tasks after exploiting the iterative pruning mechanism. However, the amount of preserved parameters  becomes larger with the continuous growth of new tasks. 
When almost all the parameters are co-used with these that are newly added in the growing step, the past parameters act like inertia given that only very few new parameters are able to be adapted freely, which potentially slows down the learning process and make the solution  non-optimal.
To alleviate this issue, we propose a task-specific Parallel Residual Function  (denoted as PRF) to increase new capacity for BERT and helps it maintain important knowledge learned from a new task. 
Specifically, we add a low-dimensional multi-head attention layer in parallel to each layer of BERT.\footnote{The idea here is a bit similar to that used in ~\cite{yuan2020parameter}, which however designs a different PRF and solves a very different research problem.} Here, we use $\mathbf{PRF}(\cdot)$ to represent the low-dimensional multi-head attention layer. 

In PRF, we reduce the dimension of the hidden state $\mathbf{h}$ from $d_m$ to a much smaller one $d_p$ by projection, which is computed as follows:
\begin{equation}
\label{eq:PRF}
\small
    {\rm \mathbf{PRF}}\left(\mathbf{h}\right) = {W_{m,p}}{\rm \mathbf{MHA}}_{p}\left({W_{p,m}}\mathbf{h}\right)
\end{equation}
where $\mathbf{MHA}_{p}$ denotes the multi-head attention layer for PRF. $W_{m,p}$ and $W_{p,m}$ are projection parameters that 
are shared across all the 12 layers in BERT without uncertainty-regularization. 
Each $d_{m}$-dimensional hidden state $\mathbf{h}$ is converted to a $d_{p}$-dimensional representation $\hat{\mathbf{h}}$, which is inputted into the multi-head attention layer ($\mathbf{MHA}_{p}$). After the multi-head attention layer,  $\hat{\mathbf{h}}$ is then converted back to the $d_{m}$-dimensional hidden state $\mathbf{h}$. 
In total, there are only about 1.6\% additional parameters are added to BERT.

\subsection{The Overall Model}
Compared with the standard BERT model, we have applied the iterative pruning with uncertainty regularization (IPR) on the linear transforms in BERT and added a low-dimensional multi-head attention layer in parallel to each layer of BERT. We define each layer of our model (denoted as $\mathbf{\rm BL}_{IPR}$) as follows: 
\begin{equation}
\label{eq:IPR}
\small
  \mathbf{\rm BL}_{IPR}\left(\mathbf{h}\right) = \mathbf{LN}\left(\mathbf{MHAL}_{\mathit{IPR}}(\mathbf{h}) + \mathbf{FFN}_{\mathit{IPR}}(\mathbf{h}) 
+ \mathbf{PRF}(\mathbf{h}) \right)
\end{equation}
where $\mathbf{PRF}(\cdot)$ is given in Eq. (\ref{eq:PRF}) without pruning and regularization. $\mathbf{MHAL}_{\mathit{IPR}}(\cdot)$ and $\mathbf{FFN}_{\mathit{IPR}}(\cdot)$ represent the multi-head attention layer (after layer normalization) and the fully connected layer of BERT with pruning and regularization. Finally, our model can be simply optimized by stochastic gradient descent (i.e., AdamW \cite{loshchilov2018fixing}) like the original BERT model.

\section{Experimental Setup}
\subsection{Datasets}
We evaluate our IPRLS approach on 16 popular datasets for sentiment classification. The first 14 datasets are product reviews collected from Amazon\footnote{https://www.amazon.com/} by \cite{blitzer2007biographies}. The IMDB \cite{maas2011learning} and MR \cite{pang2005seeing} datasets are movie reviews. Each review is classified as either positive or negative. 
Each dataset is randomly split into the training set (70\%), the development set (10\%), and the testing set (20\%).
We provide the detailed statistics of all the datasets in Table \ref{table1}. 

\begin{table}
  \centering
    \caption{Statistics of experimental datasets. Avg.L denotes the average length of reviews in each task.}
    \begin{tabular}{cccccc}
    \toprule
    \textbf{Dataset} & Train & Dev.  & Test  & Avg. L & Vocab. \\
    \midrule
    Books & 1400  & 200   & 400   & 159   & 62K \\
    Electronic & 1400  & 200   & 400   & 101   & 30K \\
    DVD   & 1400  & 200   & 400   & 173   & 69K \\
    Kitchen & 1400  & 200   & 400   & 89    & 28K \\
    Apparel & 1400  & 200   & 400   & 57    & 21K \\
    Camera & 1400  & 200   & 400   & 130   & 26K \\
    Health & 1400  & 200   & 400   & 81    & 26K \\
    Music & 1400  & 200   & 400   & 136   & 60K \\
    Toys  & 1400  & 200   & 400   & 90    & 28K \\
    Video & 1400  & 200   & 400   & 156   & 57K \\
    Baby  & 1400  & 200   & 400   & 104   & 26K \\
    Magazines  & 1400  & 200   & 400   & 117   & 30K \\
    Software & 1400  & 200   & 400   & 129   & 26K \\
    Sports & 1400  & 200   & 400   & 94    & 30K \\
    IMDB  & 1400  & 200   & 400   & 269   & 44K \\
    MR    & 1400  & 200   & 400   & 21    & 12K \\
    \bottomrule
    \end{tabular}%
  \label{table1}%
\end{table}%

\begin{table*}[htbp]
  \centering
    \caption{The classification accuracy evaluated on the final model after all 16 tasks are visited. We use Avg. to represent the average accuracy of all tasks for each method. The numbers with * indicate that the improvement of IPRLS over the corresponding baseline is statistically significant with $p<0.05$ under t-test.}  
  \scalebox{1}{
    \begin{tabular}{ccp{1.2cm}<{\centering}p{1.2cm}<{\centering}p{1.2cm}<{\centering}p{1.2cm}<{\centering}p{1.2cm}<{\centering}p{1.2cm}<{\centering}p{1.2cm}<{\centering}p{1.2cm}<{\centering}p{1.2cm}<{\centering}}
    \toprule
    Task ID & Task  & Bi-LSTM & TextCNN & BERT  & Piggyback & PackNet & UCL   & Re-init & IPRLS\\
    \midrule
    1     & Magazines & 77.72*  & 80.45*  & 59.33*  & 95.42  & 95.00  & 92.25*  & 94.33  & 95.67  \\
    2     & Apparel & 79.57*  & 83.09*  & 56.50*  & 90.50  & 89.92*  & 91.08  & 89.75*  & 91.75  \\
    3     & Health & 79.33*  & 82.13*  & 57.92*  & 94.42  & 94.17  & 92.33*  & 93.42  & 94.00  \\
    4     & Camera & 77.72*  & 83.97*  & 55.58*  & 91.67*  & 91.33*  & 91.75*  & 92.92  & 92.92  \\
    5     & Toys  & 76.36*  & 83.65*  & 80.75*  & 90.83*  & 93.17  & 92.58*  & 91.42*  & 93.17  \\
    6     & Software & 76.68*  & 83.01*  & 82.58*  & 92.75  & 93.42  & 89.25*  & 94.00  & 92.42  \\
    7     & Baby  & 76.84*  & 82.37*  & 92.08*  & 92.58  & 93.83  & 92.67  & 93.25  & 93.25  \\
    8     & Kitchen & 78.21*  & 82.29*  & 89.75  & 90.75  & 92.50  & 89.25  & 91.17  & 90.75  \\
    9     & Sports & 78.13*  & 84.29*  & 91.17*  & 91.92  & 92.58  & 92.17  & 91.75  & 92.67  \\
    10    & Electronics & 73.08*  & 80.69*  & 89.67*  & 90.58  & 93.17  & 90.17*  & 90.92  & 92.00  \\
    11    & Books & 77.64*  & 83.65*  & 91.92  & 90.75  & 89.50*  & 92.92  & 90.50*  & 92.83  \\
    12    & Video & 77.80*  & 81.01*  & 89.75  & 91.08  & 88.17*  & 90.00  & 91.33  & 91.25   \\
    13    & IMDB  & 75.16*  & 82.45*  & 89.42  & 88.75  & 87.33*  & 89.25  & 89.42  & 89.75 \\
    14    & DVD   & 80.61*  & 81.89*  & 89.50  & 87.25*  & 87.33*  & 90.17   & 87.92*  & 89.75 \\
    15    & Music & 77.08*  & 79.81*  & 89.83  & 85.50*  & 87.83*  & 88.42 & 85.50*  & 89.75  \\
    16    & MR    & 71.55*  & 72.12*  & 85.33  & 82.17*  & 80.25*  & 84.58 & 83.33  & 84.33  \\
    \midrule
          & Avg.  & 77.09*  & 81.68*  & 80.69*  & 90.43*  & 90.59*  & 90.55*  & 90.68*  & 91.64  \\
    \bottomrule
    \end{tabular}%
    }
  \label{table2}%
\end{table*}%

\subsection{Baseline Methods}
We compare our IPRLS approach with several sentiment classification and continual/lifelong learning methods. First, we compare our method with three widely-used text classification models: BiLSTM \cite{graves2005framewise} that calculates the input sequence from forward and backward directions; TextCNN \cite{kim2014convolutional} that applies a CNN with convolutional kernel size in [3,4,5]; BERT \cite{devlin2019bert} that has been proved to achieve superior performance in various NLP tasks. In addition, we also compare 
IPRLS with PackNet \cite{mallya2018packnet}, Piggyback \cite{mallya2018piggyback} and UCL \cite{ahn2019uncertainty}, which are popular continual learning methods. For fair comparison, we replace the CNNs in original PackNet, Piggyback and UCL papers with the BERT model, and choose the optimal hyperparameters based on the development sets.

\subsection{Implementation Details}
We use a pre-trained BERT$_{\rm BASE}$ model \cite{devlin2019bert} as base network. The BERT$_{\rm BASE}$ has 12 Transformer layers, 12 self-attention heads, and 768 hidden dimensions. We use the default BERT vocabulary in our experiments.
The embedding layer for the words in vocabulary is frozen during the training process. 
We set the maximum sequence length of input reviews to be 256. 
The weights of linear transforms in PRF are initialized by sampling from the normal distribution $\mathcal{N}(0,0.02)$, and the bias terms are initialized to be zero. For the iterative pruning with uncertainty regularization module, both initial network training and network re-training are trained for 3 epochs. We use the AdamW optimization algorithm \cite{loshchilov2018fixing} to optimize the whole IPRLS model.
The batch size is set to be 32. The learning rate of the parallel residual function is set to be 1e-4 for initial network training process and 1e-5 for network retraining process. The hyperparameters defined in Eq. (\ref{eq:REG}) are set to $\alpha = 0.1$, $\beta = 0.1$, $\gamma =0.03$. $L_2$ regularization (weight decay$=$4e-5) is applied to the MHA and FFN functions for avoiding overfitting. 


We also tune the hyper-parameters for all baselines with the validation sets. To ensure the reliability and stability of the experiments, we run each model three times with randomly initialized parameters and report the averaged values. 

\section{Experimental Results}
\subsection{Overall Performance Comparison}
We conduct experiments following the common lifelong/continual learning setting. The experimental data from 16 different domains arrive 
sequentially, and each dataset is considered as a separate task. We run all methods with the same task ordering during training.
The classification accuracy on the test set of each domain is reported after all 16 tasks are visited. That is, each model keeps learning a new task by using the weights learned from past tasks as initialization. To analyze the capability of our IPRLS approach in gaining knowledge from past tasks to enhance the current task (positive transfer), we also report the results of the base BERT model whose parameters are re-initialized after a task has been trained (denoted as Re-init). In particular, Re-init applies a separate model for each task and takes up much more space than our IPRLS approach. 

Table \ref{table2} demonstrates the experimental results of our IPRLS approach and baselines. From the results, we can observe that the typical deep neural models (i.e., Bi-LSTM, TextCNN, BERT) perform much worse than the lifelong/continual learning methods (i.e., Piggyback, PackNet, UCL and IPRLS), especially for the first several tasks. For example, BERT performs as well as Piggyback, PackNet, and IPRLS on the recent tasks, but significantly worse on the first several tasks. This is consistent with our main claim as conventional deep neural models do not purposely preserve old knowledge and inevitably suffer from catastrophic forgetting.

For the continual learning-based methods, Piggyback shows comparable performance to Re-init, but only the binary masks are learned for separate tasks. Although the Piggyback method can, to some extent, alleviate the catastrophic forgetting problem, the old knowledge cannot be used effectively to enhance the performance of the new tasks. 
PackNet achieves promising results on the first few tasks, but it struggles to retain performance on some latest ones. For example, the accuracy of PackNet is only 80.25\% on the last MR task, which is much worse than that of IPRLS (84.33\%). This may be because the amount of preserved parameters increases as the number of old tasks grows, and there are fewer parameters for new tasks, making the learned solution immature. On the contrary, UCL behaves excellent on the latest few tasks, but suffers from the catastrophic forgetting issue slightly on the first few tasks. 
Our IPRLS method achieves better results than the compared methods on both old tasks and new tasks. 
We believe the improvement mainly comes from the design of iterative pruning with uncertainty regularization method and the parallel residual function, which endows IPRLS the capability of resolving the so-called stability-plasticity dilemma.

\begin{figure}
    \centering
    \includegraphics[width=0.85\columnwidth]{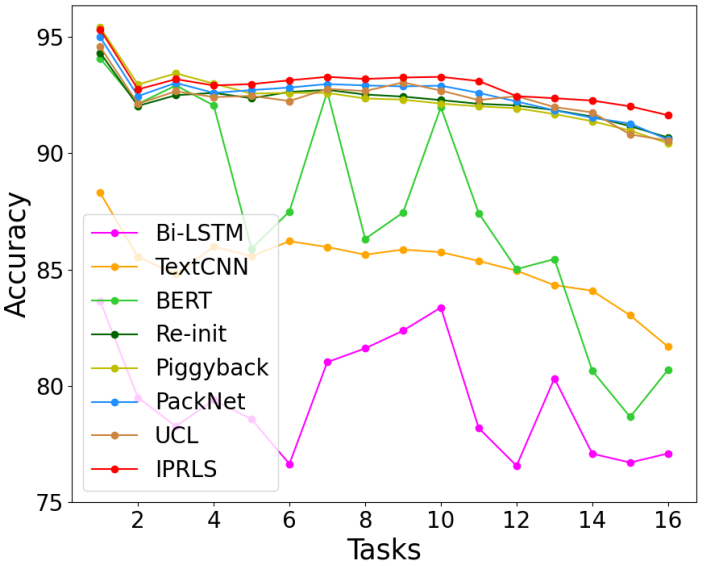}
    \caption{The average test accuracy over the $1$-st to $k$-th tasks after learning the $k$-th task ($1 \leq k \leq 16$).}
    \label{fig:all_avg_acc}
\end{figure}

\begin{figure}
    \centering
    \includegraphics[width=1.0\columnwidth]{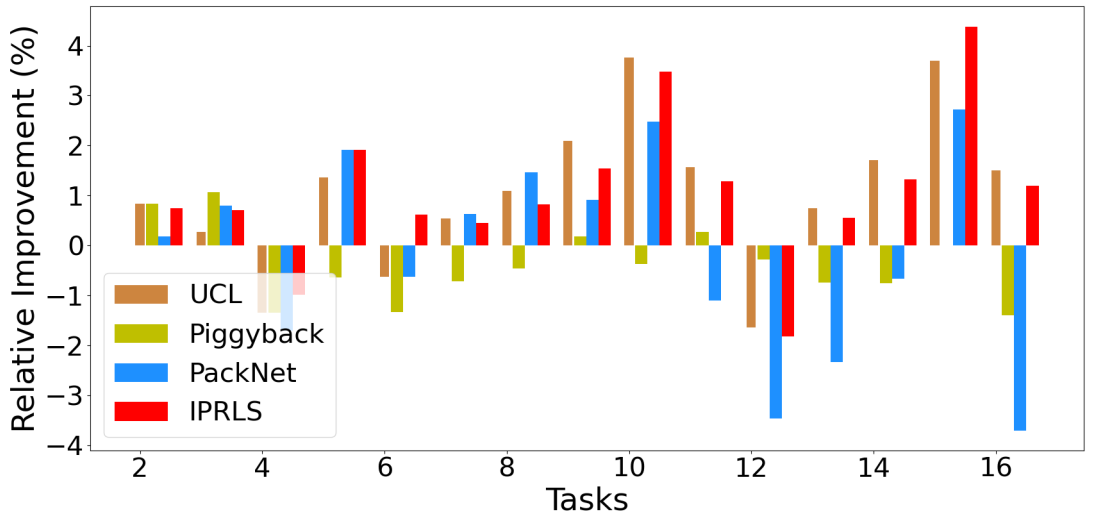}
    \caption{The relative improvement (average test accuracy) of the four continual learning methods over the Re-init model.} 
    \label{fig:forward}
\end{figure}

\subsection{Visible Analysis on Middle States}
In this section, we provide more insights into the middle states of the learned models. Figure \ref{fig:all_avg_acc} illustrates the average accuracy over the $1$-st to $k$-th tasks after learning the $k$-th task ($1 \leq k \leq 16$). 
The results allow for several interesting observations. 
First, we observe that among the standard supervised sentiment classifiers, BERT performs much better than BiLSTM and TextCNN. The improvements of BERT is mainly because of its large pre-trained model encodes more general knowledge than BiLSTM and TextCNN. However, to our surprise, the performance of BERT  drops sharply when $k$ equals to 13, even worse than BiLSTM and TextCNN. This indicates that BERT may suffer from catastrophic forgetting more severely than BiLSTM and TextCNN. In addition, BERT has larger fluctuations than BiLSTM and TextCNN. 
Second, we observe that these continual learning methods are resilient and keep relatively stable accuracy in their whole learning stages. In particular, IPRLS exhibits better accuracy than PackNet and UCL on new tasks and old tasks, respectively. 



\begin{figure}[h]
    \centering
    \includegraphics[width=0.85\columnwidth]{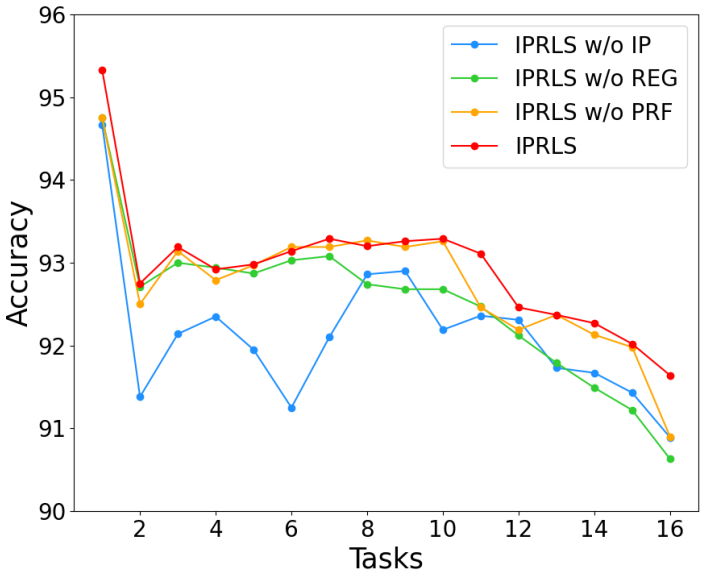}
    \caption{The average test accuracy of learned tasks for ablation test.}
    \label{fig:albation}
\end{figure}

\begin{figure}[h]
    \centering
    \includegraphics[width=0.85\columnwidth]{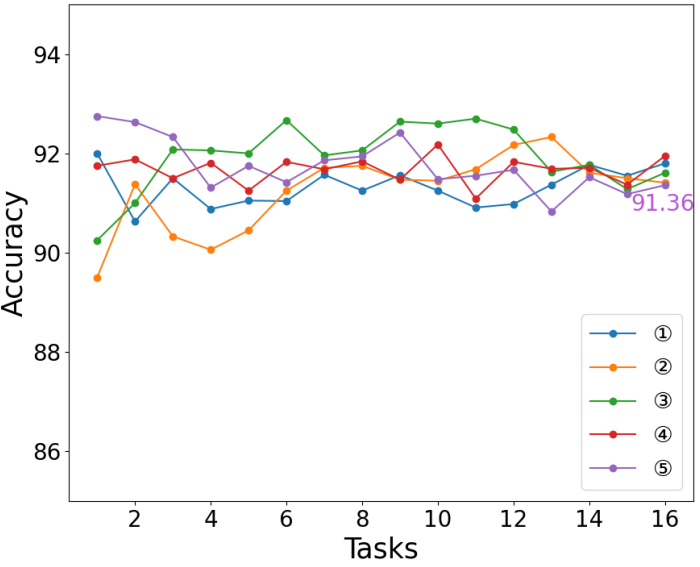}
    \caption{Experiment results of IPRLS with 5 different orderings randomly sampled from the 16 domains.}
    \label{fig:shuffle}
\end{figure}

\begin{figure}[!h]
    \centering
    \includegraphics[width=1\columnwidth]{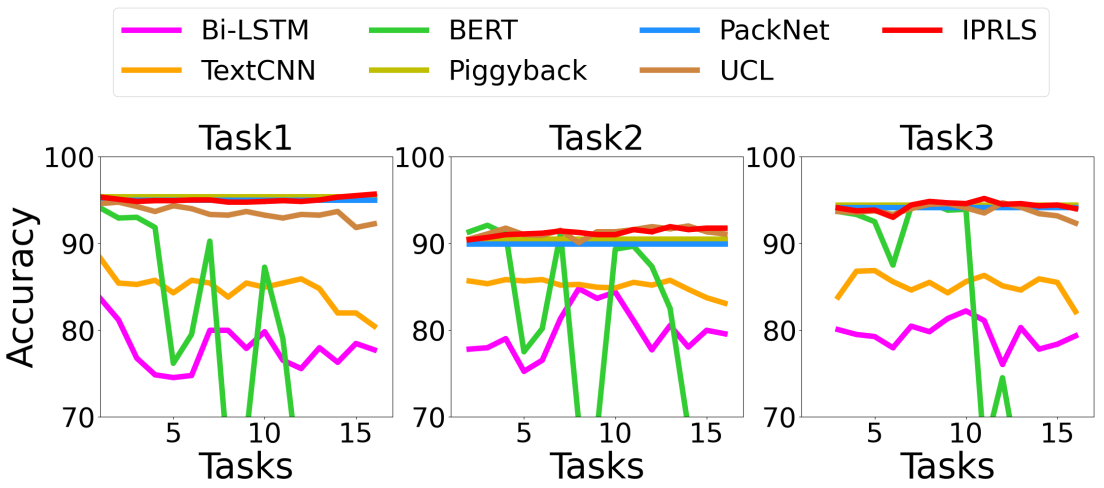}
    \caption{The change of test accuracy for the first 3 tasks during the whole learning process.}
    \label{fig:tasks_init5}
\end{figure}

\subsection{Assessing Backward Knowledge Transfer}
In Figure \ref{fig:tasks_init5}, we show how the results of tasks 1-3 change as subsequent new tasks are being learned subsequently. Taking the first task as an example, we observe that the accuracy of the BERT model starts to decay sharply after learning 5 tasks, probably because the knowledge learned from these new tasks interferes with what was learned previously. However, IPRLS achieves stable accuracy over the whole learning process by using the iterative pruning with uncertainty regularization technique to resolve the catastrophic forgetting issue. 
In addition, the results of IPRLS (especially for task 2) increase slightly while learning more new tasks. This verifies that IPRLS enables positive backward transfer,  i.e. learning new tasks improves performance on past tasks.



\subsection{Assessing Forward Knowledge Transfer}
We also assess the capability of  IPRLS  to use the knowledge gained from past tasks to help learn the new task. Specifically, we assess the forward knowledge transfer of the lifelong/continual learning methods by training these models on task 1 to $k+1$ and testing them on task $k+1$ ($1 \leq k \leq 15$). Figure \ref{fig:forward} reports the relative improvements of the lifelong learning methods over the Re-init method for tasks 2-16.  
From Figure \ref{fig:forward}, we can observe that PackNet struggles to retain forward transfer as an 
increase of the number of tasks. 
For most tasks, Piggyback obtain negative forward knowledge transfer. 
Our IPRLS method shows a promising ability to both forward and backward knowledge transfer for lifelong sentiment classification. 

\subsection{Ablation Study}
We conduct ablation study to investigate the effectiveness of each component in IPRLS. 
First, we discard the iterative pruning strategy, and merely adopt the uncertainty regularization and the parallel residual function to perform lifelong learning (denoted as IPRLS w/o IP). Second, we remove the uncertainty regularization strategy from IPRLS, denoted as IPRLS w/o REG. Third, to analyze the contribution of the parallel residual module for sentiment analysis, we remove the parallel residual function from the BERT framework (denoted as IPRLS w/o PRF). 
Figure \ref{fig:albation} reports the average test accuracy of the learned tasks for the ablation test.  

From the results in Figure \ref{fig:albation}, we can make the following observations. 
First, the iterative pruning method has the largest impact on the performance of IPRLS. The performance of IPRLS drops shapely when discarding the iterative pruning, and the drops are particularly significant for tasks 2-8. This is because
the iterative pruning strategy enables the model to release redundant weights for learning new tasks without affecting the performance of previous tasks, especially when the number of new tasks is small.
Second, the performance of IPRLS  drops gradually when discarding the uncertainty regularization, and the drops are particularly significant for tasks 7-16. This is reasonable since the uncertainty regularization strategy plays 
a critical role in mitigating the catastrophic forgetting problem when the number of new tasks is large.
Since the amount of preserved parameters is proportional to the number of tasks, the learning process of new tasks would slow down without regularization, making the learned solution immature. 
Third, the parallel residual function also makes a contribution to IPRLS, while the improvement of parallel residual function is smaller than iterative pruning and uncertainty regularization. The results of IPRLS w/o PRF show similar trends with that of IPRLS. 

\subsection{Effect of Task Ordering}
In this section, we explore the effect of task ordering in lifelong sentiment classification. Since there are a large number of different task orderings, it is difficult, if not impossible, to investigate all task orderings. In this experiment, we randomly sample 5 different task orderings (task sequences). Figure \ref{fig:shuffle} reports the average test accuracy of different tasks  with three independent runs per ordering. From the results, we can observe that although IPRLS presents kinds of different accuracy with different task orderings, it is in general insensitive to orders because the results are quite close and almost show the similar trends, especially for the last 3 tasks. 

\subsection{Error Analysis}
To better understand the limitations of IPRLS, we additionally carry out a careful analysis of the errors made by the IPRLS model after all the 16 tasks are trained sequentially. In particular, we select 150 instances that are incorrectly predicted by IPRLS from the 16 datasets.
Generally, we reveal several reasons for the classification errors which can be divided into the following categories. 
(1) IPRLS fails to classify some documents that are lengthy and contain more than 256 tokens. This is because the long documents often contain complex structures and complicated sentiment expressions. 
(2) The second error category occurs 
when the input documents contain  implicit opinion
expressions or require deep comprehension. This may be because the training data 
is not sufficient enough such that IPRLS cannot capture the latent patterns. It suggests that a certain stronger comprehension strategy needs to be devised in the future so as to capture latent features better.
(3) The third error category is caused by the opposite sentiment polarities contained in the document. It is even difficult for humans to decide the sentiment of the document. 
(4) The fourth most common error category includes examples with comparative opinions. The documents typically
express a comparative opinion on two or more entities in terms of the shared features or attributes. It is difficult to identify the preferred entity in each comparative document.

\section{Conclusion and Future Work}
In this paper, we presented an iterative pruning with uncertainty regularization method for improving the performance of lifelong sentiment classification.  A low-dimensional parallel residual function was added in parallel to each layer of BERT, which helps to learn better task-specific knowledge with only a small fraction of additional parameters. We conducted comprehensive experiments on 16 sentiment classification domains. Experimental results demonstrated that IPRLS performed significantly better than strong competitors.
In the future, we would like to devise a policy network to determine the pruning ratio automatically and adaptively. In addition, we also plan to design a more advanced network expansion method to further improve the performance on new tasks.  

\begin{acks}
This work was partially supported by National Natural Science Foundation of China (No. 61906185, 92046003, 61976204, U1811461), Natural Science Foundation of Guangdong Province of China (No. 2019A1515011705), Youth Innovation Promotion Association of CAS China (No. 2020357), Shenzhen Science and Technology Innovation Program (Grant No. KQTD20190929172835662), Shenzhen Basic Research Foundation (No. JCYJ20200109113441941).  
\end{acks}   

\balance
\bibliographystyle{ACM-Reference-Format}
\bibliography{acmart}


\end{document}